\documentclass[10pt,twocolumn,letterpaper]{article}

\usepackage{iccv}
\usepackage{authblk}
\usepackage{times}
\usepackage{epsfig}
\usepackage{graphicx}
\usepackage{amsmath}
\usepackage{amssymb}
\usepackage{enumitem}
\usepackage{multirow}
\usepackage{booktabs}
\usepackage{marvosym}
\usepackage{indentfirst} 
\usepackage{graphicx}
\usepackage{bbm}
\usepackage[T1]{fontenc}
\usepackage[utf8]{inputenc}
\usepackage[english]{babel}
\usepackage[autostyle, english=american]{csquotes}
\usepackage{subcaption}
\usepackage{ulem}

\MakeOuterQuote{"}
\usepackage{bbm}
\usepackage[breaklinks=true,bookmarks=false]{hyperref}

\iccvfinalcopy % *** Uncomment this line for the final submission

\def\iccvPaperID{****} % *** Enter the ICCV Paper ID here

\ificcvfinal\fi
\ificcvfinal\fi
\begin{document}
\normalem
%%%%%%%%% TITLE
\title{Solution for Emotion
Prediction Competition of Workshop on Emotionally and Culturally Intelligent AI}
\author{

% Authors
Shengdong Xu $^1$,
Zhouyang Chi$^1$
% Yang Yang$^1$,
 Yang Yang\thanks{Corresponding author: Yang Yang(yyang@njust.edu.cn)} $^1$,
}

\affil{ 
 $^1$Nanjing University of Science and Technology
 % $^2$Dalian University of Technology
 % $^3$Tsinghua University 
}

\maketitle
\setlength{\intextsep}{1pt}
\setlength{\abovecaptionskip}{1.5pt}
\begin{abstract}

 This report provide a detailed description of the method that we explored and proposed in the WECIA Emotion Prediction Competition (EPC), which predicts a person's emotion through an artistic work with a comment. The dataset of this competition is ArtELingo, designed to encourage work on diversity across languages and cultures. The dataset has two main challenges, namely modal imbalance problem and language-cultural differences problem. In order to address this issue, we propose a simple yet effective approach called single-multi modal with Emotion-Cultural specific prompt(ECSP), which focuses on using the single modal message to enhance the performance of multimodal models and a well-designed prompt to reduce cultural differences problem. To clarify, our approach contains two main blocks: (1)XLM-R\cite{conneau2019unsupervised} based unimodal model and X$^2$-VLM\cite{zeng2022x} based multimodal model (2) Emotion-Cultural specific prompt. Our approach ranked first in the final test with a score of 0.627.
\end{abstract}
%%%%%%%%%%%%%%%%%%%%%%%%%%%%%%%%%%%%%%%%%%%%%%%%%%%%%%%
\begin{figure*}
    \centering
    \includegraphics[scale=0.49]{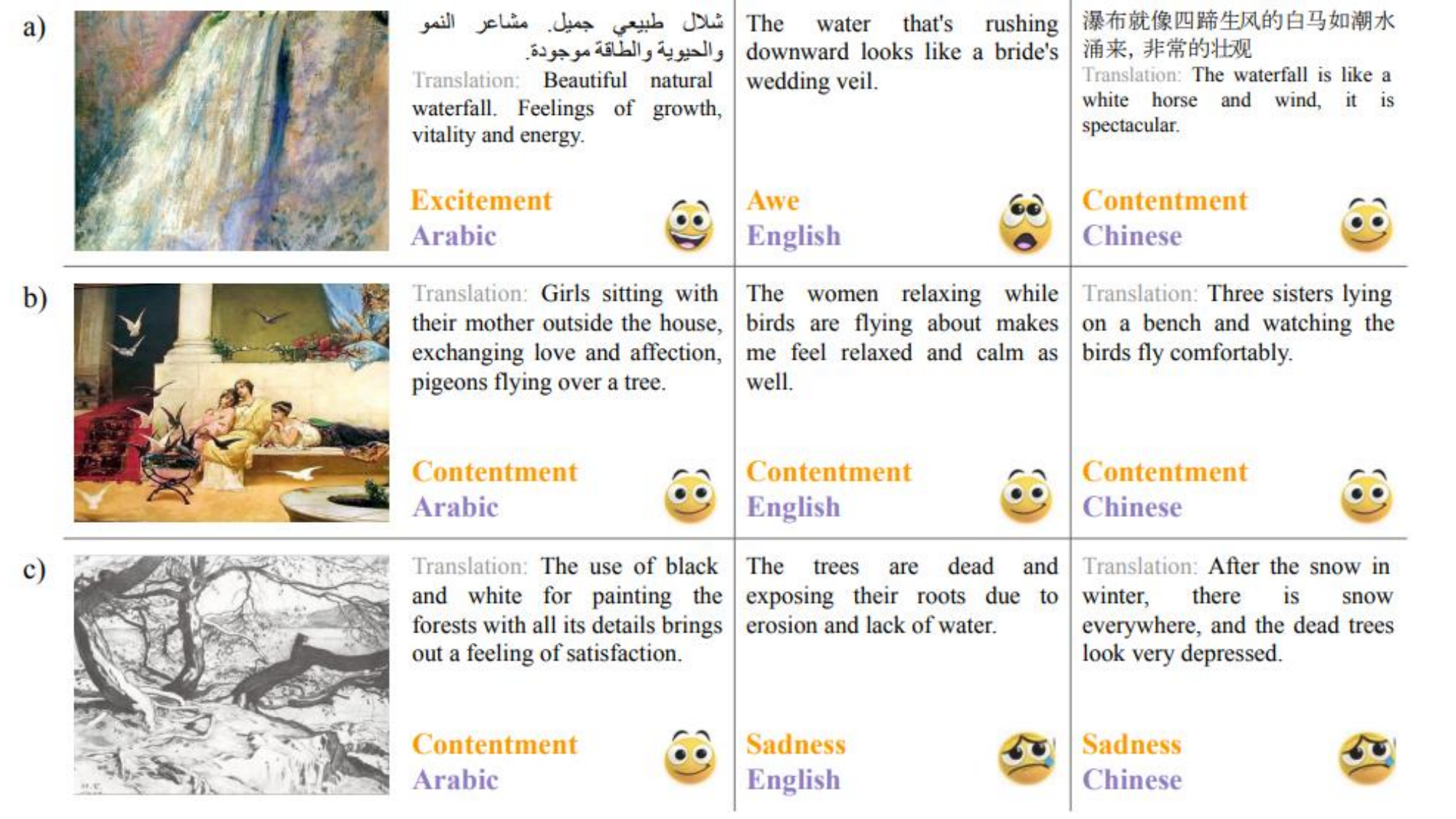}
    \caption{ArtELingo, a mulitlingual dataset and benchmark of WikiArt with captions \&emotions.}
    \label{fig: dataset}
    \vspace{-10pt}
\end{figure*}
%%%%%%%%%%%%%%%%%%%%%%%%%%%%%%%%%%%%%%%%%%%%%%%%%%%%%%%

%%%%%%%%% BODY TEXT
\section{Introduction}

In recent years, multimodal learning\cite{yang2019comprehensive, yang2019semi, yang2019semi_2} has gradually become one of the research hotspots in the fields of machine learning and data mining. It has been successfully applied in various real-world scenarios, such as multimedia search\cite{yang2021rethinking}, multilingual processing, and image captioning\cite{yang2022exploiting}. Traditional emotion prediction methods primarily utilize a single modality of text. However, this competition focuses on emotion classification of image-text pairs.\par

Emotion prediction plays a critical role in text classification for tasks such as stance detection and lie detection\cite{yang2023deep}. As shown in Figure \ref{fig: dataset} , ArtElingo\cite{mohamed2022artelingo} dataset compares and contrasts annotations on WikiArt across language/culture. These differences are interesting and important. One might suggest using machine translation to translate English captions to many other languages, but this will miss much of the opportunity. So we should build human-compatible AI that is more aware of our emotional being is important for increasing the social acceptance of AI. In this competition, we used a language model and a vision-language pretrain model: XLM-R and X$^2$-VLM as our baseline.\par
% In this competition, we adopted a zero-shot strategy and explored three recent approaches: OmniMotion\cite{abs-2306-05422}, TAPIR\cite{abs-2306-08637}, and Cotraker\cite{abs-2307-07635}. Among them, OmniMotion and Cotraker had poor generalization on the Perception Test, while TAPIR performed relatively better and was used as our baseline.
%%%%%%%%%%%%%%%%%%%%%%%%%%%%%%%%%%%%%%%%%%%%%%%%%%%%%%%
%数据分析以及问题提出
Consider Figure 1c, where an Arabic annotator assigned the image the label \textbf{contentment}, but the other two annotators used the label : \textbf{sadness}. Captions are useful for diving deeper into these differences. The \textbf{sadness} annotations mention death and disasters, in contrast with the contentment annotation that ends with : felling of satisfaction. Further, the Table \ref{tab: baseline} show the method provided by organizer. Obviously, using only one of these modes can't achieve good results.The Vit can't even surpass the accuracy of guessing, which is 1/9. The reason is already mentioned above. Therefore, we can find that language modal has more rich information and should be the main modal of training. The reason for the poor performance of BERT is also very obvious. BERT model didn't consider multilingualism and cultural differences issues.\par
\begin{figure*}
    \centering
    \includegraphics[page=1,scale=0.49]{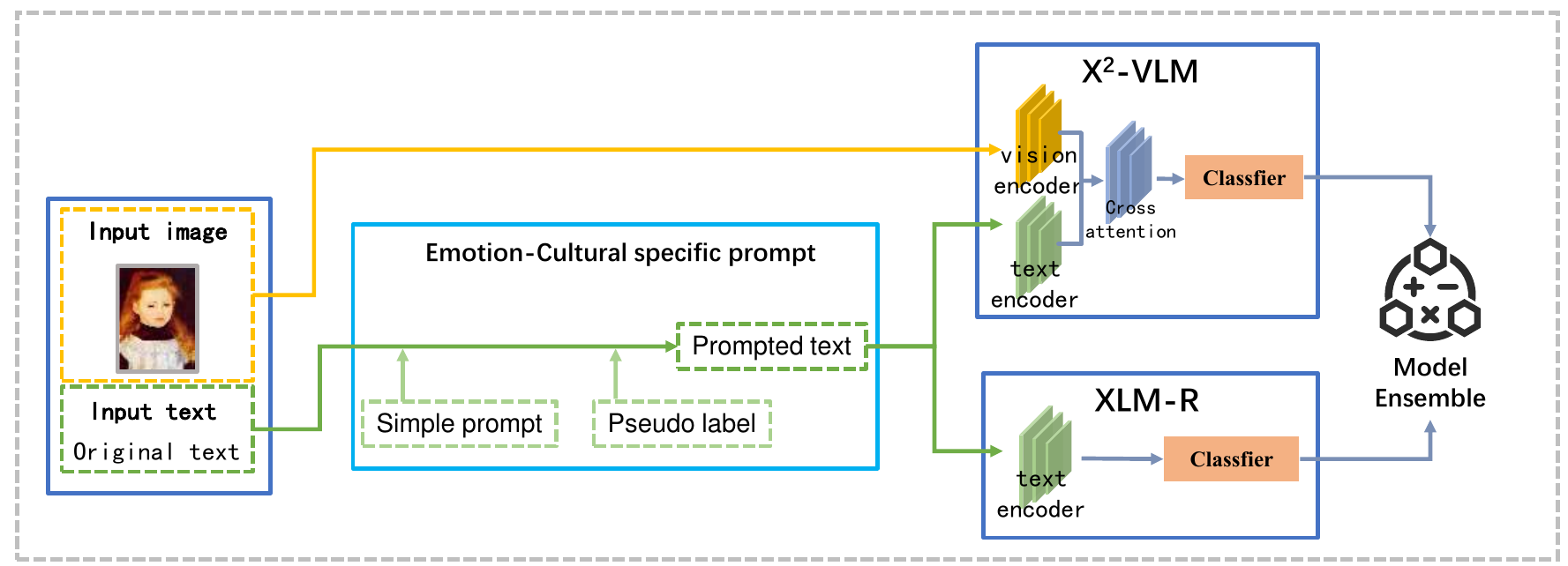}
    % \vspace{-55pt}
    \caption{Outline of our proposed method. Pseudo Labels
Retrieval Module is shown in the figure \ref{fig: PLRM }. The construction details of Emotion-Cultural specific prompted text will be introduced below.}
    \label{fig: architecture}
    \vspace{-10pt}
\end{figure*}

\begin{table}[htp]
    \centering
    \begin{tabular}{cccccc}
    \toprule
    Method & F1 & ACC \\
    \hline
    BERT & 0.56  & 0.68  \\
    Vit & 0.01 & 0.00 \\
    \toprule
    \end{tabular}
\caption{The result of single modal baseline provided by organizer.}
\label{tab: baseline}
\end{table}
% By analyzing TAPIR's motion trajectory predictions, we have identified two issues, as illustrated in Figure \ref{fig: problem} (b): \textbf{a) Static Point Jitter:} In static camera videos, points occasionally showed slight positional jitter despite being static in the Ground Truth Tracker. \textbf{b) Static Point Pseudo-Following:} While predicting point trajectories, originally static points often shifted due to the coverage of moving objects, either following the movement of the objects or remaining at a position deviating from the initial point after some time frames of moving along with the objects.
%%%%%%%%%%%%%%%%%%%%%%%%%%%%%%%%%%%%%%%%%%%%%%%%%%%%%%%
%解决方法
To address these challenges, we propose single-multi modal with Emotion-Cultural specific prompt, as shown in Figure \ref{fig: architecture}. Firstly, in order to improve the ability of mining information from language modal, we chose two different model. One from single modal and one from multimodal. By using the way of model ensemble, we hope different mining way will have a complementary effect. Secondly, we meticulously design a language prompt which based on the retrieval augmented method. Retrieval augmented method help us find the nearest sample in the training dataset. Thus, we can get a pseudo-label and construct our prompt.

%我们的贡献
We introduce single-multi modal with Emotion-Cultural specific prompt strategy, and its contributions can be summarized as follows:
\begin{itemize}
[itemsep=0pt,parsep=0pt,topsep=0pt,partopsep=0pt,leftmargin=*]
    \setlength{\itemindent}{1.3em}
    \item We use the strategy of integrating single modal and multimodal models to enhance the mining of more informative language modalities.
    \item We design a Emotion-Cultural specific prompt method which based on the retrieval augmentation\cite{yang2024alignment, yang2015auxiliary}. It solved part of cultural differences problem.
\end{itemize}

%方法
%base model
%prompt
\section{Method}

\subsection{Base Model}
In our work, we take XLM-R and X$^2$-VLM as our base model. 
% In our work, we take TAPIR as our base model. It employs a two-stage approach, involving matching and refinement, to independently query and predict fine-grained point trajectories and features based on local information. Unlike previous algorithms, TAPIR measures uncertainty in a self-supervised manner to estimate point trajectories with greater precision. It starts by globally comparing query point features with features from other frames to establish an initial tracking estimate. Subsequently, local features are extracted from a neighborhood around the initial estimate and compared to query features at a higher resolution. A temporal depth convolution network is then used for post-processing to refine the similarity, resulting in more accurate position estimates. TAPIR's methodology significantly enhances the precision of point trajectory estimation.
\vspace{-10pt}
\subsubsection{XLM-R} 
\setlength{\parindent}{2em}  XLM-R is a transformer-based \cite{devlin2018bert} language model trained with the multilingual MLM objective on 100 languages, three languages in artelingo dataset included. In order to deal with multi-language issues, XLM-R proposed new methods for data processing and model optimization objectives. The former uses Sentence Piece with a unigram language model to build a shared sub-word vocabulary, and the latter introduces a supervised optimization objective of translation language modeling(TLM). In this competition, we directly added a linear layer to fine-tune the pre-trained model for the task of emotion prediction. \par
\vspace{-10pt}
\subsubsection{X$^2$-VLM}

Obviously, the task of competition is a multimodal emotion classification task. So we chose a vision-language pre-training model. X$^2$-VLM, an all-in-one model with flexible modular architecture, in which further unify imgae-text pre-training and video-text pre-training in one model. So it's able to learn unlimited visual concepts associated with diverse text descriptions. But the origin X$^2$-VLM model can't deal with multilingual issues. Therefore, we replaced the origin text encoder(BERT) with the XLM-R model mentioned before. 
\subsection{Emotion-Cultural specific prompt}
In order to alleviate the modal imbalance\cite{yang2016learning} problem mentioned before and mine the potential information of text modality, we focus on designing a prompt \cite{dhamyal2022describing}based on emotion and culture. We proposed the Emotion-Cultural specific prompt. It consists of two parts, namely simple prompt and prompt with retrieval augmentation\cite{guu2020retrieval}.
\vspace{-10pt}
\subsubsection{Simple prompt}
First, we build a simple prompt using the attributes in the dataset. Given a annotation, denoted as $ann$, it contains the following attributes: \textbf{art\textunderscore style, language, utterance}, the first two of which provide some information about the image. Then we splice them as follows:  

\textbf{\textit{The art style of image is \{art\textunderscore style\}. There is a comment from a \{language\} person. What emotions did he express? amusement, awe, contentment, excitement, anger, disgust, fear, sadness or something else,\{utterance\}.}}  

Although, maybe the vision encoder and text encoder of X$2$-VLM can learn the art style of image and the language information of text. But we found that adding this information can effectively improve the performance of the model. What's more, adding the category information to the input text can help the model better understand the goals of the task.
\vspace{-10pt}
\subsubsection{Prompt with retrieval augmentation}
To alleviate cultural differences issues, we use labels of samples who share the same language with the current sample to construct pseudo-labels and embed them into the simple prompt constructed before. The specific plan is as follows: Assume that the current sample is $s_i$, and the embedding of its text and image are: $v_{embed_{i}}$, $t_{embed_{i}}$, which are obtained through text and image encoders (e.g. CLIP) respectively, and then splice them together to get $Embed_{i}:$ [$v_{embed_{i}}$, $t_{embed_{i}}$], and calculate the similarity with samples of the same language (cultural differences in the same language will be smaller), sort the similarities from large to small, retrieve the top-k samples(in this competition, we just chose the most similar sample which means the k equals one), and splice the labels of these samples as pseudo labels and embed them into prompt. The formulas to select pseudo label are as follows:
\begin{equation}
    \begin{split}
       &\ \textbf{index}^{i} = {\underset{j=1,...,n}{\arg\max} \, (Sim(\textbf{Embed}_{i},\textbf{Embed}_{j}))} \\&
    \end{split}
\end{equation}
\vspace{-20pt}
\begin{equation}
    \begin{split}
        &\ \textbf{pseudo-label}^{i} = \textbf{label}^{\textbf{index}^{i}}\  \\&
        \textbf{if}\ Sim(\textbf{Embed}_{i},\textbf{Embed}_{\textbf{index}^{i}}) > \eta 
    \end{split}
\end{equation}
where we chose cosine similarity as the $Sim$ function, and $\eta$ is the hyperparameter, indicating the threshold of cosine similarity. The final prompt we constructed is as follows: 

\textbf{\{Simple prompt\}. \{Input text\}. The emotion this picture is most likely trying to express is \{pseudo label\}}.

\begin{figure*}
    \centering
    \includegraphics[page=2,scale=0.49]{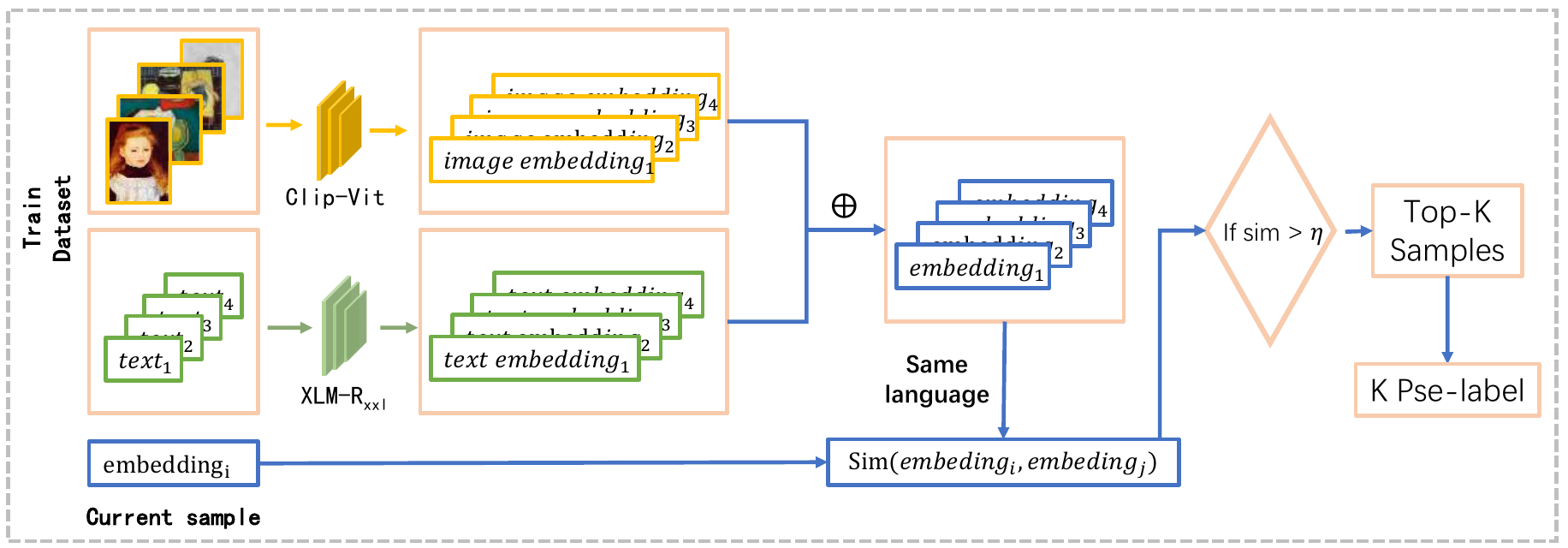}
    % \vspace{-55pt}
    \caption{Pseudo Labels
Retrieval Module. First, use the text encoder and image encoder \cite{radford2021learning}  to get the embeddings of text and images, and then Concatenate them together. Then use cosine similarity to calculate the embeddings of the current sample's embedding and other samples, determine whether it exceeds the threshold, and take the labels of the top-k samples as pseudo labels.}
    \label{fig: PLRM }
    \vspace{-10pt}
\end{figure*}
\vspace{-10pt}
\subsection{Test Time Augmentation}
\begin{figure}
    \centering    
    \includegraphics[width=1\columnwidth]{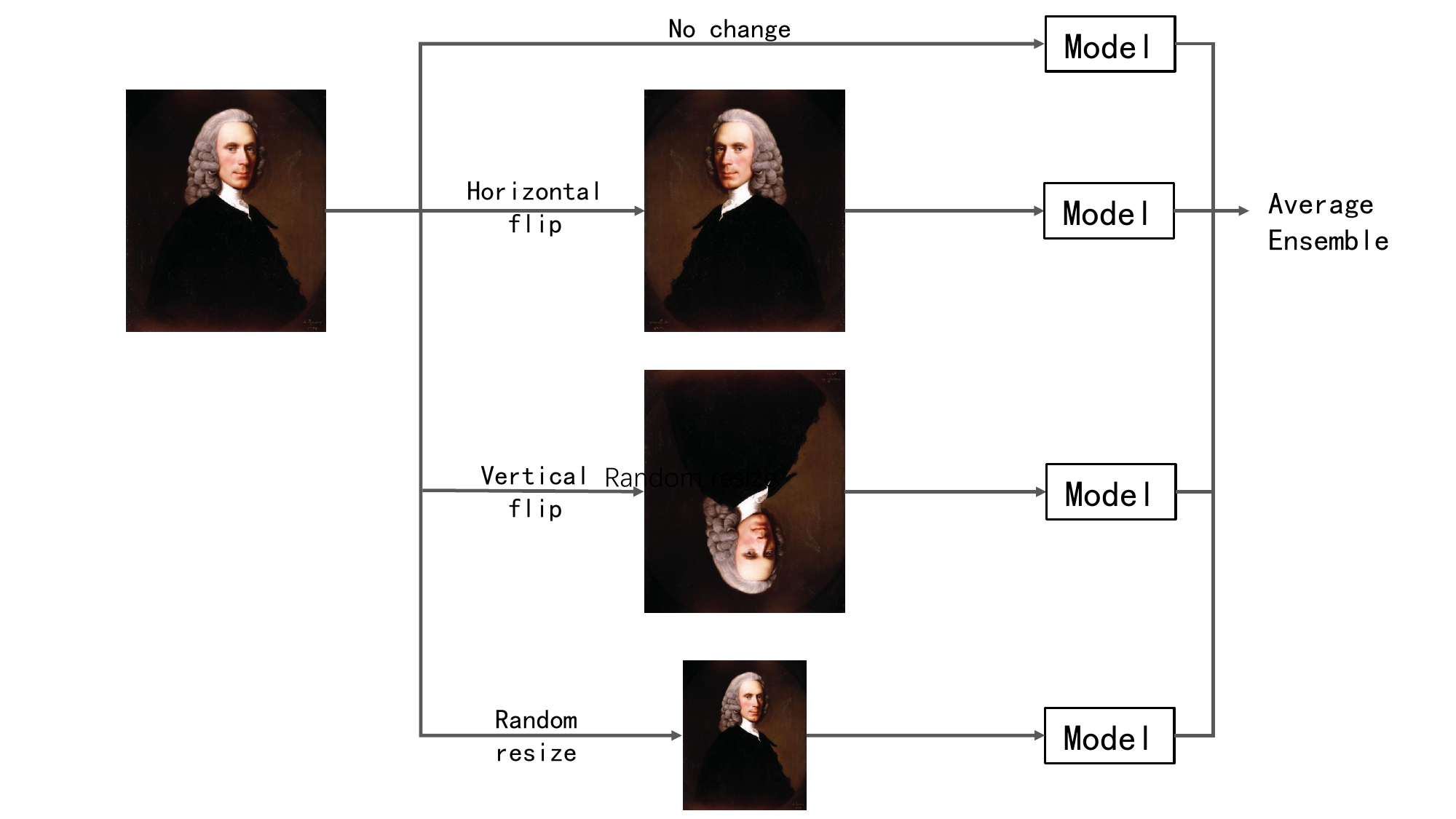}
    \caption{Test Time Augmentation: The original image is input into the model with text through four transformations}
    \label{fig: tta}    
    \vspace{-15pt}
\end{figure}

Considering that the effect of the image modality is not as good as the text modality, and to further improve the robustness of the multi-modal model X$^2$-VLM, we use test-time augmentation enhancement technique in the inference stage which contains horizontal flipping, vertical flipping and random cropping of images, as shown in the figure \ref{fig: tta}. The following Experiments prove that it has a certain effect on the final results.
\subsection{Discussion}
Although prompt learning has a certain enhancement effect on multimodal classification, constructing a reasonable and robust prompt itself poses great challenges. Moreover, in this competition, the constructed prompt serves as generic input and cannot personalize the learning well. Addressing this issue remains an intriguing and open research challenge. Future work may build more efficient prompts to deal with the above problems.

\section{Experiment}
\textbf{Dataset.} Our method is fine-tuned on the pre-trained models of XLM-R and X$^2$-VLM using official datasets. But We split the training and validation sets randomly instead of following the official way.

% Our approach, as a zero-shot method, exclusively relies on the pre-trained model and weights provided by TAPIR, with no additional data used for training. Concerning the pre-trained data, TAPIR made modifications to the publicly available MOVi-E dataset generation script to create a new dataset named MOVi-F, tailored for training data. Specifically, adjustments were made to the camera angles in the script, ensuring that the 'look at' point followed a linear trajectory near the bottom of the workspace, traveling through the center of the workspace. Both the validation and test sets utilize the official data provided.

\textbf{Metric.} The evaluation metric for this task is the F1 Score, which is the harmonic mean of Precision and Recall.

\textbf{Implementation Detail.} 
The maximum number of text tokens used by the single-modal method is 90. The maximum number of text tokens for multi-modal model text is 100. The resolution of the image is 768×768. Our approach is founded on fine-tune principles and makes use of the pre-trained model made available on the official XLM-R \footnote{\label{myfootnote1}$https://huggingface.co/xlm-roberta-base/tree/main$} and X$^2$-VLM website
\footnote{\label{myfootnote2}$https://lf-robot-opensource.bytetos.com/obj/lab-robot-public/x2vlm_ckpts_2release/x2vlm_base_1b.th$}. Regarding hyperparameter settings, we configure the similarity threshold as $\eta =0.75$ 

\begin{table}[htp]
\vspace{1.0em}
    \centering
    \begin{tabular}{cccccc}
    \toprule
    Method & F1 & Acc \\
    \hline
    $\textbf{BERT}_{baseline*}$ & 0.56  & 0.68  \\
    $\textbf{Vit}_{baseline*}$ & 0.01  & 0.00  \\
    $\textbf{XLM-R}_{large}$ & 0.613 & 0.725 \\
    $\textbf{XLM-R}_{large}$+ecsp & 0.618 & 0.729 \\
    \textbf{X$^2$-VLM} &  0.619 & 0.730  \\
    \textbf{X$^2$-VLM}+ecsp & 0.622 & 0.735  \\
    \textbf{Model Ensemble}  & \textbf{0.627} &  \textbf{0.741}  \\
    \toprule
    \end{tabular}
\caption{Comparison method. *: The experimental data is sourced from official authorities. +prompt refers to the effect of adding the best prompt, as shown in the table \ref{tab: abl}.}
\label{tab: compare}
\vspace{-1.0em}
\end{table}

\textbf{Comparison Methods Result.} Table \ref{tab: compare} shows the F1 score performance, from which we can observe that: 1) Vit exhibits poor performance in the competition dataset, because there is a situation in the data set where a image corresponds to text with different emotions, as shown in figure \ref{fig: dataset}. 2) X$^2$-VLM outperforms baseline methods by achieving about 0.059 and 0.609 performance improvements respectively. This phenomenon exhibits that text modality plays a far more important role in emotion classification than image modality.
Note that the single-modal model is the large version, but the multi-modal text encoder is the base version, but the effects of the two are very close, indicating that the image modality still has a positive effect on the results.

\textbf{Ablation Study.} To analyze the contribution of the prompt in our method, we conduct more ablation studies in this competition. The F1 score after adding different prompts is demonstrated in Table \ref{tab: abl}. From the table, we can clearly see that after adding prompt, the F1 value has a certain improvement, whether in single mode or multi-mode. At the same time, for the same model, after adding escp, the effect gain is greater than adding sp, which proves the effectiveness of the Emotion-Cultural specific prompt we designed. The results indicate that prompts contributes to improvement, and better prompts can lead to further enhancements.
\begin{table}[htp]
\vspace{1.0em}
    \centering
    \begin{tabular}{cccccc}
    \toprule
    Method & F1 & Acc \\
    \hline
    $\textbf{XLM-R}_{large}$ & 0.613 & 0.725 \\
    $\textbf{XLM-R}_{large}$+sp & 0.614 & 0.728 \\
    $\textbf{XLM-R}_{large}$+pseudo-label & 0.612 & 0.724 \\
    $\textbf{XLM-R}_{large}$+ecsp & 0.618 & 0.729 \\
    \textbf{
    X$^2$-VLM} &  0.619 & 0.730  \\
    \textbf{X$^2$-VLM}+sp & 0.621 & 0.733  \\
    \textbf{X$^2$-VLM}+pseudo-label & 0.620 & 0.731  \\
    \textbf{X$^2$-VLM}+ecsp & 0.622 & 0.735  \\
    \toprule
    \end{tabular}
\caption{Ablation experiment. sp and escp denote simple prompt and Emotion-Cultural specific prompt respectively.} 
\label{tab: abl}
\end{table}

\section{Conclusion}

This report summarized our solution for the Emotion Recognition Challenge in the Workshop on Emotionally and Culturally Intelligent AI. Our approach was based on the Emotion-Cultural specific prompt and Test Time Augmentation, thereby effectively enhancing the overall performance of emotion recognition.

{\small
\bibliographystyle{ieee_fullname}
\bibliography{Emotion_Prediction}

\begin{thebibliography}{10}\itemsep=-1pt

\bibitem{conneau2019unsupervised}
Alexis Conneau, Kartikay Khandelwal, Naman Goyal, Vishrav Chaudhary, Guillaume Wenzek, Francisco Guzm{\'a}n, Edouard Grave, Myle Ott, Luke Zettlemoyer, and Veselin Stoyanov.
\newblock Unsupervised cross-lingual representation learning at scale.
\newblock {\em arXiv preprint arXiv:1911.02116}, 2019.

\bibitem{devlin2018bert}
Jacob Devlin, Ming-Wei Chang, Kenton Lee, and Kristina Toutanova.
\newblock Bert: Pre-training of deep bidirectional transformers for language understanding.
\newblock {\em arXiv preprint arXiv:1810.04805}, 2018.

\bibitem{dhamyal2022describing}
Hira Dhamyal, Benjamin Elizalde, Soham Deshmukh, Huaming Wang, Bhiksha Raj, and Rita Singh.
\newblock Describing emotions with acoustic property prompts for speech emotion recognition.
\newblock {\em arXiv preprint arXiv:2211.07737}, 2022.

\bibitem{guu2020retrieval}
Kelvin Guu, Kenton Lee, Zora Tung, Panupong Pasupat, and Mingwei Chang.
\newblock Retrieval augmented language model pre-training.
\newblock In {\em International conference on machine learning}, pages 3929--3938. PMLR, 2020.

\bibitem{mohamed2022artelingo}
Youssef Mohamed, Mohamed Abdelfattah, Shyma Alhuwaider, Feifan Li, Xiangliang Zhang, Kenneth~Ward Church, and Mohamed Elhoseiny.
\newblock Artelingo: A million emotion annotations of wikiart with emphasis on diversity over language and culture.
\newblock {\em arXiv preprint arXiv:2211.10780}, 2022.

\bibitem{radford2021learning}
Alec Radford, Jong~Wook Kim, Chris Hallacy, Aditya Ramesh, Gabriel Goh, Sandhini Agarwal, Girish Sastry, Amanda Askell, Pamela Mishkin, Jack Clark, et~al.
\newblock Learning transferable visual models from natural language supervision.
\newblock In {\em International conference on machine learning}, pages 8748--8763. PMLR, 2021.

\bibitem{yang2023deep}
Yang Yang, Ran Bao, Weili Guo, De-Chuan Zhan, Yilong Yin, and Jian Yang.
\newblock Deep visual-linguistic fusion network considering cross-modal inconsistency for rumor detection.
\newblock {\em Science China Information Sciences}, 66(12):222102, 2023.

\bibitem{yang2019semi_2}
Yang Yang, Zhao-Yang Fu, De-Chuan Zhan, Zhi-Bin Liu, and Yuan Jiang.
\newblock Semi-supervised multi-modal multi-instance multi-label deep network with optimal transport.
\newblock {\em IEEE Transactions on Knowledge and Data Engineering}, 33(2):696--709, 2019.

\bibitem{yang2024alignment}
Yang Yang, Jinyi Guo, Guangyu Li, Lanyu Li, Wenjie Li, and Jian Yang.
\newblock Alignment efficient image-sentence retrieval considering transferable cross-modal representation learning.
\newblock {\em Frontiers of Computer Science}, 18(1):181335, 2024.

\bibitem{yang2019comprehensive}
Yang Yang, Ke-Tao Wang, De-Chuan Zhan, Hui Xiong, and Yuan Jiang.
\newblock Comprehensive semi-supervised multi-modal learning.
\newblock In {\em IJCAI}, pages 4092--4098, 2019.

\bibitem{yang2022exploiting}
Yang Yang, Hongchen Wei, Hengshu Zhu, Dianhai Yu, Hui Xiong, and Jian Yang.
\newblock Exploiting cross-modal prediction and relation consistency for semisupervised image captioning.
\newblock {\em IEEE Transactions on Cybernetics}, 2022.

\bibitem{yang2015auxiliary}
Yang Yang, Han-Jia Ye, De-Chuan Zhan, and Yuan Jiang.
\newblock Auxiliary information regularized machine for multiple modality feature learning.
\newblock In {\em Twenty-Fourth International Joint Conference on Artificial Intelligence}, 2015.

\bibitem{yang2016learning}
Yang Yang, De-Chuan Zhan, and Yuan Jiang.
\newblock Learning by actively querying strong modal features.
\newblock In {\em IJCAI}, pages 2280--2286, 2016.

\bibitem{yang2019semi}
Yang Yang, De-Chuan Zhan, Yi-Feng Wu, Zhi-Bin Liu, Hui Xiong, and Yuan Jiang.
\newblock Semi-supervised multi-modal clustering and classification with incomplete modalities.
\newblock {\em IEEE Transactions on Knowledge and Data Engineering}, 33(2):682--695, 2019.

\bibitem{yang2021rethinking}
Yang Yang, Chubing Zhang, Yi-Chu Xu, Dianhai Yu, De-Chuan Zhan, and Jian Yang.
\newblock Rethinking label-wise cross-modal retrieval from a semantic sharing perspective.
\newblock In {\em IJCAI}, pages 3300--3306, 2021.

\bibitem{zeng2022x}
Yan Zeng, Xinsong Zhang, Hang Li, Jiawei Wang, Jipeng Zhang, and Wangchunshu Zhou.
\newblock X$^{2}$-vlm: All-in-one pre-trained model for vision-language tasks.
\newblock {\em arXiv preprint arXiv:2211.12402}, 2022.

\end{thebibliography}
}
\end{document}